\documentclass[letterpaper, 10 pt, conference]{ieeeconf}  %
\IEEEoverridecommandlockouts                              %
\usepackage{graphics}    %
\usepackage{times}       %
\usepackage{amsmath}     %
\usepackage{amssymb}     %
\usepackage{graphicx}
\usepackage{algorithm}
\usepackage[noend]{algpseudocode}
\usepackage{glossaries}
\usepackage[table]{xcolor}
\usepackage{multirow} 
\usepackage{booktabs}
\usepackage{subcaption} %
\usepackage{balance}

\usepackage{amsopn}

\newcommand{\bT}{\mathbf{T}}
\newcommand{\bw}{\mathbf{w}}
\newcommand{\bW}{\mathbf{W}}

\newcommand{\cS}{\mathcal{S}}
\newcommand{\cC}{\mathcal{C}}

\newcommand{\bbR}{\mathbb{R}}

\newcommand{\bn}{\mathbf{n}}

\newcommand{\bp}{\mathbf{p}}

\newcommand{\bbSE}{\mathbb{SE}}

\newcommand{\bmu}{\boldsymbol{\mu}}

\newcommand{\bLambda}{\mathbf{\Lambda}}

\def\g2o{$g^2o$}
\def\se3{\mathrm{SE}(3)}
\def\t2v{\mathrm{t2v}}
\def\v2t{\mathrm{v2t}}

\def\secref#1{Sec.~\ref{#1}}
\def\figref#1{Fig.~\ref{#1}}
\def\tabref#1{Tab.~\ref{#1}}
\def\eqref#1{Eq.~(\ref{#1})}
\def\algref#1{Alg.~\ref{#1}}

\def\ie{{i.e.}}

\def\lidar{LiDAR}
\def\lidars{LiDARs}

\newacronym{slam}{SLAM}{Simultaneous Localization and Mapping}
\newacronym{sfm}{SfM}{Structure from Motion}
\newacronym{ba}{BA}{Bundle Adjustment}
\newacronym{sdf}{SDF}{Signed Distance Function}
\newacronym{pgo}{PGO}{Pose-Graph Optimization}
\newacronym{vpr}{VPR}{Visual Place Recognition}
\newacronym{sgd}{SGD}{Stochastic Gradient Descent}
\newacronym{ils}{ILS}{Iterative Least-Squares}
\newacronym{icp}{ICP}{Iterative Closest Point}
\newacronym{gn}{GN}{Gauss-Newton}
\newacronym{lm}{LM}{Levenberg-Marquardt}
\newacronym{pcg}{PCG}{Preconditioned Conjugate Gradient}
\newacronym{map}{MAP}{Maximum-A-Posteriori}
\newacronym{gf}{GF}{Gaussian Filters}
\newacronym{pf}{PF}{Particle Filters}
\newacronym{sdp}{SDP}{Semi-Definite Programming}
\newacronym{bst}{BST}{Binary Search Tree}
\newacronym{ndt}{NDT}{Normal Distributed Transform}
\newacronym{pca}{PCA}{Principal Component Analysis}
\newacronym{lo}{LO}{LiDAR odometry}
\newacronym{agv}{AGV}{Autonomous Ground Vehicle}
\newacronym{vbr}{VBR}{A Vision Benchmark in Rome}
\newacronym{nc}{NC}{Newer College Dataset}
\newacronym{ate}{ATE}{Absolute Trajectory Error}

\title{\LARGE \bf MAD-BA: 3D LiDAR Bundle Adjustment -- from Uncertainty Modelling to Structure Optimization}
\author{
	Krzysztof \'{C}wian \quad\quad
	Luca Di Giammarino \quad\quad
	Simone Ferrari \quad\quad
	Thomas Ciarfuglia\\\\
	\centering
	 \quad\quad
        Giorgio Grisetti \quad\quad
	Piotr Skrzypczyński
\thanks{Work of K. \'Cwian was supported by the Polish National Agency for Academic Exchange under the STER program ``Towards Internationalization of Poznan University of Technology Doctoral School'' (2022-2024).}
\thanks{This work was partially supported by funds from the FESR Lazio 2021-2027 Program - ``Riposizionamento competitivo RS'' (DE n. G18823, 28/12/2022). CUP: B83C23003970007 - Project: ``Digital Twin Cit''.}
\thanks{Krzysztof \'{C}wian and Piotr Skrzypczyński are with the Institute of Robotics and Machine Intelligence, Poznan University of Technology, Poznan, Poland, Email:\,\,{\tt\footnotesize{\{krzysztof.cwian, piotr.skrzypczynski\}@put.poznan.pl.}}}
\thanks{Luca Di Giammarino, Simone Ferrari, Thomas Ciarfuglia and Giorgio Grisetti are with the Department of Computer, Control, and Management Engineering ``Antonio Ruberti", Sapienza University of Rome, Italy,
Email:\,\,{\tt\footnotesize{\{digiammarino, s.ferrari, ciarfuglia, grisetti\}@diag.uniroma1.it.}}}%
\thanks{This work has been supported by PNRR MUR project PE0000013-FAIR and by PUT project 0214/SBAD/0252.}
\thanks{Accepted version of the manuscript. The final published version is available at IEEE Xplore: https://doi.org/10.1109/LRA.2025.3573628}
}

\begin{document}
\maketitle
\thispagestyle{empty}
\pagestyle{empty}

\begin{figure*}[t]
  \centering
 \includegraphics[width=0.99\linewidth]{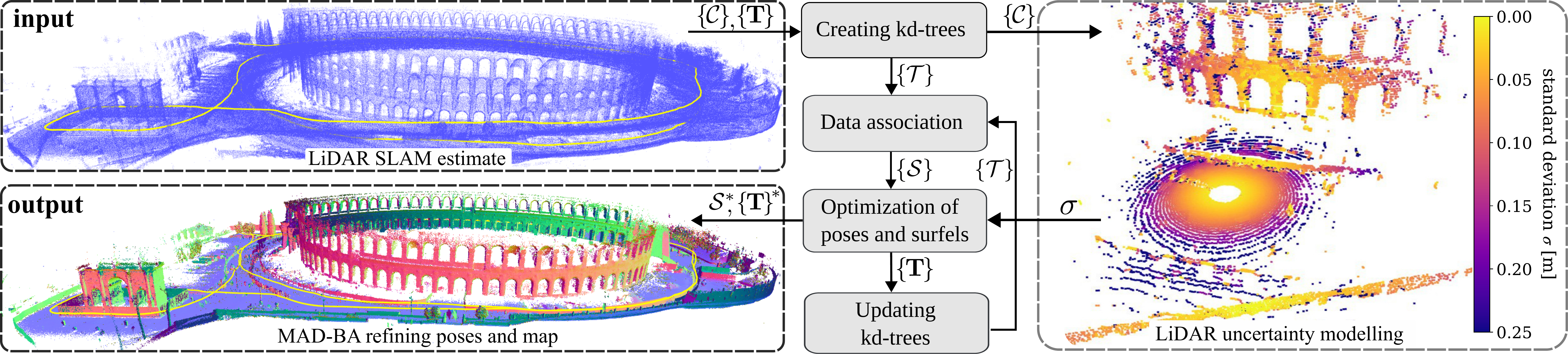}
 \caption{\textbf{Pipeline.} Given a trajectory estimated by a \lidar{} SLAM system  our method jointly optimizes the poses and map. It leverages on our uncertainty model to weight the optimization. In MAD-BA map, colors represent normal vectors of surfels.}
  \label{fig:motivation}
  \vspace{-4mm}
\end{figure*}

\begin{abstract}

The joint optimization of sensor poses and 3D structure is fundamental for state estimation in robotics and related fields. Current \lidar{} systems often prioritize pose optimization, with structure refinement either omitted or treated separately using implicit representations. 
This paper introduces a framework for simultaneous optimization of sensor poses and 3D map, represented as surfels. A generalized \lidar{} uncertainty model is proposed to address less reliable measurements in varying scenarios. Experimental results on public datasets demonstrate improved performance over most comparable state-of-the-art methods. The system is provided as open-source software to support further research.
\end{abstract}

\section{Introduction}
\label{sec:intro}

Accurate and consistent 3D modeling is essential for applications such as autonomous navigation, robotics, AR/VR, infrastructure inspection, and environmental monitoring. Modern 3D \lidar{} sensors capture millions of points per second, enabling mapping of large-scale environments. Achieving globally consistent maps requires precise alignment of scans into a unified representation, which must be accurate, scalable, and resilient to noise while accommodating dynamic environments and incomplete data.

\gls{ba} is the gold standard in 3D reconstruction due to its ability to jointly optimize sensor poses and structure for globally consistent results \cite{Triggs1999}. This technique is well-suited for visual SLAM, where dense and structured image data provide rich visual cues for feature extraction and correspondence matching \cite{engels2006bundle}. In contrast, \lidar{} systems face unique challenges, including sparse and irregular data distributions and the lack of visual cues, complicating feature extraction and efficient representation. Conventional \lidar{} \gls{slam} techniques often focus on trajectory refinement via \gls{pgo} \cite{Behley2018, di2022md}, which keeps a fixed structure of the map during optimization, resulting in uncorrected noise and inconsistencies in point clouds \cite{zhang2016degeneracy}. 

Recent methods have sought to address these challenges. Filtering approaches using implicit representations like \gls{sdf} \cite{Deng2023} and hierarchical optimization methods \cite{liu2023hba} refine trajectories but treat pose and structure optimization separately, which can lead to inconsistencies. Photometric refinement techniques adapted from vision \cite{di2023photometric} and robust local geometric methods like MAD-ICP \cite{ferrari2024mad} demonstrate improved robustness by leveraging full point clouds but remain limited to local registration without jointly optimizing global structure and trajectory.

This paper addresses these gaps by introducing a unified global \gls{ba} framework for \lidar{} data that integrates pose and structure optimization within an uncertainty-aware approach to achieve consistent and robust mapping.

The remainder of this paper is structured as follows. Section~\ref{sec:related} reviews related work. Section~\ref{sec:main} describes our proposed framework in detail. Section~\ref{sec:exp} presents experimental results across various datasets. Finally, Section~\ref{sec:conclusion} concludes the paper and outlines future directions.

\section{Related Work}
\label{sec:related}

Global refinement techniques, such as \gls{ba}, are extensively used in Visual SLAM and \gls{sfm} systems but remain less common for \lidar{}. Unlike images with dense and structured grids, \lidar{} data consists of sparse and irregular point clouds, particularly at longer ranges, making the application of \gls{ba} computationally challenging. Consequently, many \lidar{} frameworks emphasize local registration techniques, such as \gls{icp} and its variants, which rely on subsampled point clouds \cite{velas2016collar, vizzo2023kiss}, salient geometric features \cite{zhang2014loam}, or \gls{ndt} \cite{stoyanov2012fast}. However, while subsampling can reduce noise, it often sacrifices critical geometric details essential for accurate registration.

Classic methods like LOAM \cite{zhang2014loam} extract features such as surfaces and corners to compute point-to-plane and point-to-line distances. Recent methods, such as MAD-ICP \cite{ferrari2024mad}, demonstrate that leveraging complete point clouds without subsampling can enhance registration accuracy. MAD-ICP achieves this through robust data association using kd-tree for efficient point cloud alignment. The kd-tree-based data association is also employed in our proposed framework. SuMa \cite{Behley2018}, on the other hand, employs surfel-based map representations for efficient projective data association and real-time SLAM in large-scale environments. Both approaches primarily focus on local registration and do not perform joint pose and structure optimization, as proposed in this work. High-level geometric features such as planar segments have also been explored within a global factor graph framework to jointly estimate trajectories and maps \cite{Cwian2021}, but this approach struggles with representing accurately more complicated environments using large planar patches. 

\gls{pgo} remains a predominant approach for refining trajectories in \lidar{} \gls{slam} \cite{di2022md, Behley2018}. By representing trajectories as graphs of poses linked by relative transformations, \gls{pgo} achieves computational efficiency but fixes pairwise measurements during \gls{slam}, which can lead to drift and inconsistencies over time \cite{zhang2016degeneracy}. Hierarchical methods, such as HBA \cite{liu2023hba}, reduce computational complexity but depend on accurate initial trajectories, limiting robustness in challenging conditions.

For global refinement, Liu and Zhang \cite{liu2021balm} proposed a trajectory optimization framework using LOAM-style features, while \cite{di2023photometric} adapted photometric refinement methods from vision to high-resolution \lidar{} data. However, these approaches struggle with robustness and scalability in lower-resolution or noisy conditions. Decoupled methods, such as \gls{sdf} \cite{lan2025vdb} and implicit mapping \cite{Deng2023}, aim to filter noise while compactly representing the environment but fail to integrate pose and structure refinement into a unified framework.

\lidar{} measurements are inherently noisy due to various environmental conditions. Modeling range uncertainty allows reliable measurements to be prioritized, preventing optimization degradation and ensuring effective contributions to \gls{ba} \cite{grisetti2020least}. However, existing uncertainty models for 2D scanners \cite{Skrzypczynski2007} or airborne LiDARs \cite{wehr1999airborne} lack generalizability across diverse \lidar{} sensors and application scenarios.

To address these challenges, we propose a unified global optimization framework that jointly refines \lidar{} poses and structure using surfel-based map representations. By incorporating a generalized \lidar{} uncertainty model, our approach mitigates noise and sparsity, achieving improved global consistency and robustness. Our contributions include:
\begin{enumerate}
	\item a scalable and robust \gls{ba} framework tailored to \lidar{} data;
	\item a generalized \lidar{} uncertainty model for weighting range measurements;
	\item open-source implementation for reproducibility and community use.
\end{enumerate}

\section{MAD-BA}
\label{sec:main}

This section introduces our novel approach to 3D \lidar{} Bundle Adjustment, which jointly optimizes scene representations and sensor poses to ensure geometric consistency and accuracy (\figref{fig:motivation}). The method directly processes raw 3D point cloud data using surfels -- compact, disk-shaped primitives that efficiently capture surface geometry. The following subsections detail the methodology: surfel-based scene representation, efficient data association via kd-trees, a cost function minimizing geometric errors leveraging on a \lidar{} uncertainty model, and finally, the optimization routine using the \gls{lm} algorithm to refine poses and surfels jointly.

\subsection{Data Representation}
We represent the scene geometry using dense surfels and poses obtained from a \lidar~odometry or a SLAM pipeline. A pose is parameterized as a 3D homogeneous transformation matrix $\bT_k \in \se3$.

A surfel $s = \left< \bn_s, \bp_s, r_s \right>$ in our method is an oriented disc characterized by a center point $\bp_s \in \bbR^3$, a surface normal vector $\bn_s \in \bbR^3$, and a radius $r_s \in \bbR$. Surfels are well-suited for representing the scene due to their efficiency in fusion and updates during BA and their adaptability to structural changes like loop closures. Compared to voxel-based approaches, surfels can more effectively capture thin surfaces and detailed geometry. Unlike mesh-based methods, surfels do not require computationally expensive updates to the topology during optimization.

\subsection{Data Association}
\label{sec:data}
Similarly to Ferrari \emph{et al.} \cite{ferrari2024mad}, to match estimated surfels with \lidar{} measurements, we employ kd-trees \cite{ferrari2024mad}. For each new cloud $\mathcal{C}_k$ delivered by the \lidar, a kd-tree $\mathcal{T}_k$ is built. This preprocess results in a data structure that encodes plane segmentation of the cloud and allows nearest-neighbor queries.
A generic leaf $l = \left < \bp_l, \bn_l \right >$ of $\mathcal{T}_k$ will contain a small subset of the point cloud. The leaf of a kd-tree is defined by the mean point $\bp_l \in \bbR^3$ and the surface normal $\bn_l \in \bbR^3$. The task is to match all the terminal nodes against the existing surfel set $\mathcal{S}$. A new surfel is created if a leaf cannot match any surfel. A leaf $l$ to be matched needs to meet the following distance criteria:
\begin{itemize}
\item the Euclidean distance between the surfel point $\bp_s$ and the leaf mean $\bp_l$ should be lower than $d_e$;
\item the Euclidean distance along the normal of the surfel $\bn_s$ should be lower than $d_{n}$;
\item the angle between normals $\bn_l$ and $\bn_{p}$ should be lower than $d_{\theta}$.
\end{itemize}
Whenever a new measurement leaf is associated with a surfel $s$, its mean, normal, and radius are updated (\secref{subsec:creation}, \secref{subsec:normal}). Leveraging the kd-tree \cite{ferrari2024mad}, nearest-neighbor queries are correct and complete, ensuring the best surfel-leaf matches. 
Thanks to the property exhibited by this kd-tree, data re-association can be efficiently performed several times during \gls{ba}. Indeed, the \gls{pca} tree-building process is the optimal choice since it leads to the minimum \textit{depth} \cite{mcnames2001fast}. \algref{alg:data-ass} complements this section. 

\begin{algorithm}
\small
\caption{Data Association}
\textbf{input}: kd-trees $\left\{ \mathcal{T} \right\}$ \\
\textbf{output}:  surfel set $\mathcal{S}$ \\
\textbf{local}: list of leaves associations $\mathcal{L}$
\begin{algorithmic}

\For{$\mathcal{T}_i$ $\in$ $\left\{ \mathcal{T}\right\}$}
    \For{$\mathcal{T}_j$ $\in$ $\left\{ \mathcal{T} \right\}$ such that $\mathcal{T}_i \neq \mathcal{T}_j$}
        \For{$l_i$ $\in$ $\mathcal{T}_i$}
            \State $\mathcal{A}_i$ $\leftarrow$ $\{\}$ \Comment{empty set of leaves associations}
            \If{$\exists \mathcal{A} \in \mathcal{L}$ such that $l_i \in \mathcal{A}$}
                \State $\mathcal{A}_i$ $\leftarrow$ $\mathcal{A}$
            \Else
                \State $\mathcal{A}_i$.add($l_i$) \Comment{leaf does not match any surfel}
                \State $\mathcal{L}$.append($\mathcal{A}_i$)
            \EndIf
            \\
            \State $l_j$ $\leftarrow$ $\mathcal{T}_j$.nearestNeighbor($l_i$)
            \If{$\mathcal{A}_i$.checkMatch($l_j$)}
                \State $\mathcal{A}_i$.add$(l_j)$ \Comment{leaf-surfel match}
            \EndIf
        \EndFor
    \EndFor
\EndFor \\

\State $\mathcal{S}$ $\leftarrow$ createSurfels($\mathcal{L}$) \Comment{surfel creation/surfel normal update}
\State \Return{$\mathcal{S}$}
\end{algorithmic}
\label{alg:data-ass}
\end{algorithm}

\subsection{Cost Function}
Our goal is to optimize surfels and poses to maximize geometrical consistency. Each pose corresponds to a 3D \lidar{} point cloud. Let $\mathcal{K}$ be the set of all poses and $\mathcal{S}_k$ be the set of all surfels with corresponding measurements in pose $k \in \mathcal{K}$. To robustly handle outliers, the residual $e$ is weighted using the Huber robust loss function $\rho_\mathrm{Huber}$ with parameter $\rho_\mathrm{ker}$. Employing the standard deviation $\sigma$ of the \lidar{} measurement, the total negative log likelihood can be compactly written as: 
\begin{equation}
E(\mathcal{K}, \mathcal{S}) = \sum_{k \in \mathcal{K}} \sum_{s \in \mathcal{S}_k}   e \left ( k, s \right ) 
\label{eq:cost-outer}
\end{equation}
The error $e(k, s)$ represents a sum of the point-to-plane distance between mean leaf $\bp_l$ and $\bp_s$ along the normal $\bn_s$ expressed in the local reference frame $\bT_w^k$:
\begin{equation}
e(k, s) = \sum_{i \in \mathcal{A}_s} \rho_\mathrm{Huber} \left ( \sigma^{-1} (\bT_w^k \bn_s) ^T \left( \bT_w^k \bp_s   - \bp_{l_i} \right) \right )
\label{eq:cost-inner}
\end{equation}
Here, $w$ is the global frame, while $k$ is the local. $\mathcal{A}_s$ is the set of leaf matches with the surfel $s$, calculated as described in \secref{sec:data}. 

\subsubsection{\lidar{} uncertainty model}
\label{sec:lidar_uncertainty}

The waveform of a \lidar{} return is influenced by sensor parameters such as beam divergence and measurement conditions like range and incidence angle. As these increase, the beam's intersection surface expands, broadening the returned waveform.
The proprietary nature of \lidars{} firmware means that the way waveforms are processed to compute ranges remains undisclosed. While the peak likely plays a central role, details on interpreting broad waveforms are unknown. Furthermore, commercial \lidar{} interfaces provide only processed point clouds, omitting raw waveform data, which limits error analysis and modeling.
Beam divergence is crucial in laser applications. In airborne laser scanning, long-range measurements yield footprints covering multiple objects in the vertical profile \cite{baltsavias1999airborne, wehr1999airborne}. Unrealistic \lidar{} simulations hinder model generalization. A physically motivated sensing model for synthesizing realistic scans is proposed in \cite{huang2023neural}.
In \cite{xu2016correction}, an optimal beam shaping method corrects intensity data, while \cite{hu2021analytical} provides an analytical formula linking waveform characteristics to measured targets. However, none offer a general uncertainty estimation. In \cite{laconte2019lidar}, the authors use external measurement tools to quantify and model LiDAR distortion coefficients. However, this approach requires careful calibration for each individual LiDAR sensor in controlled, ad-hoc environments, which limits its generalization capability.

\begin{figure}[t]
  \centering
 \includegraphics[width=0.99\columnwidth]{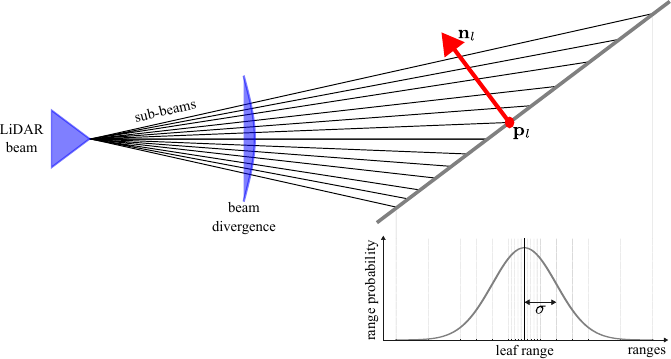}
  \caption{\textbf{\lidar{} uncertainty model.} In the top image, a single \lidar{} beam is simulated by casting a set of sub-beams towards a leaf $l$. At the bottom, the uncertainty of this measurement is modeled as a Gaussian distribution computed from the sampled ranges.}
  \label{fig:uncertainty}
  \vspace{-4mm}
\end{figure}

In contrast, we propose a sample-based method to model distortion effects and guide the optimization process, offering a more general and adaptable solution tailored for \lidar{} data. Our general approach requires only the beam divergence, typically provided in \lidar{} documentation. Using optical simulation, we cast a cone-shaped beam towards each leaf’s mean, originating from the sensor. The waveform represents the beam's echo intensity over time, discretized by casting multiple lines inside the cone (top of \figref{fig:uncertainty}). This process applies to each leaf during kd-tree construction. Although the waveform exists in time, our procedure samples a set of $N$ ranges $r_i$. The standard deviation $\sigma$ is computed from \eqref{eq:uncertainty} (bottom of \figref{fig:uncertainty}), normalized, and employed in \eqref{eq:cost-inner}.

\begin{equation}
\sigma = \sqrt{\frac{1}{N} \sum_{i=1}^N (r_i - \lVert \bp_l \rVert)^2} 
\label{eq:uncertainty}
\end{equation}

\subsection{Optimization}
We perform \gls{ba} to optimize both surfels and poses jointly. To carry on the minimization of \eqref{eq:cost-outer}, we employ the second-order \gls{lm} method. Our optimization scheme performs a number of iterations up to a maximum or until convergence. We jointly optimize both surfels and poses within each iteration by leveraging an efficient Least Squares factor graph solver \cite{grisetti2020least}. Each step is detailed in \algref{alg:mad-ba}. While all 6-dof of poses are updated, surfels are constrained to move just along their normal direction. This approach preserves local geometry, simplifies optimization by reducing degrees of freedom, and aligns with \lidar{} measurement uncertainty, which is greatest along the surfel normal. It also mitigates ambiguities in data association and ensures stable, physically meaningful refinements. 

\subsubsection{Surfel creation}
\label{subsec:creation}
Surfels are created before optimization and maintained at each iteration. This ensures high-quality data association even if the initial set of poses is not so close to the optimum. It consists of two steps: finding associations between nearest-neighbor leaves by matching each kd-tree against every other and creating/updating surfels based on these associations. The idea is to group similar leaves observed from different poses. Surfels are generated by averaging the surfel centers $\bp_s$ with the means of their corresponding leaf points $\bp_l$, and by averaging the surfel normals $\bn_s$ with the normals of the leaf points $\bn_l$. The radius $r_s$ is set as the largest radius among the corresponding leaves to ensure it fully covers the variability of the points. Considering the sparsity of a \lidar{}, this guarantees that the surfel adequately represents the local geometry, avoiding gaps or insufficient coverage that could lead to inconsistencies in the mapping process.

\subsubsection{Surfel normal updates}
\label{subsec:normal}
Measurement normals are a byproduct of the kd-tree creation, which occurs through \gls{pca} \cite{ferrari2024mad}, only once at the beginning of \gls{ba}.
The construction is a recursive procedure that involves computing mean $\bmu$ and covariance $\bLambda$ of the cloud $\mathcal{C}$. Then the eigenvectors $\bW = \begin{bmatrix} \bw_0 & \bw_1 & \bw_2 \end{bmatrix}$ of $\bLambda$ are sorted by eigenvalue. $\bw_2$ is the direction of maximum variation, while $\bw_0$ is the node's normal. For terminal nodes, the latter vector corresponds to the observed surface normal. Surfel normal update is disconnected from non-linear \gls{ba} optimization. Indeed, after each iteration, the tree leaves are re-grouped to update the surfel normal.
\subsubsection{Surfel position optimization}

After updating normals $\bn_s$ through kd-tree data matching, we jointly optimize poses and surfels using \gls{lm}. Restricting surfel movement to their normal directions $q$ allows us to parameterize the updated surfel position as $q\bn_s + \bp_s$. This joint optimization of poses and surfels imposes fewer constraints, reducing discontinuities when refining different entities. Although independent surfel optimization is faster and parallelizable \cite{schops2019bad}, the joint approach delivers superior results by effectively capturing the mutual dependencies between poses and structure, resulting in more consistent and accurate refinements.

\subsubsection{Pose optimization}
Pose updates $\delta$ are parametrized as local updates in the Lie algebra $\mathfrak{se}(3)$. Thus, the transformation $\bT_k^w$ of the local reference frame expressed in the global reference frame is updated as $\bT_k^w \mathrm{exp}(\delta).$ Local updates ensure that rotation updates are well-defined \cite{engels2006bundle}. This procedure is carried out for all the poses in the factor graph.

\begin{algorithm}
\small
\caption{MAD-BA}
\textbf{input}: clouds $\left\{\cC\right\}$, initial poses $\left\{\bT\right\}$ \\
\textbf{output}: surfel set $\mathcal{S}^*$, refined poses $\left\{\bT\right\}^*$
\begin{algorithmic}
\State $\left\{\mathcal{T}\right\}$ $\leftarrow$ buildKdTrees($\left\{ \cC \right\}$) \Comment{construct kd-trees}
\For{$i\in$ \# iterations}
    \State $\left\{\mathcal{T}\right\}$ $\leftarrow$ transform($\left\{ \mathcal{T} \right\},\left\{ \bT \right\} $) \Comment{transform kd-trees}
    \State $\cS \leftarrow$ createMatches($\left\{ \mathcal{T} \right\}$) 
    \Comment{\algref{alg:data-ass}}
    \State $\cS^*, \{\bT\}^* \leftarrow$ optimize($\cS, \{\bT\}$) \Comment{optimize poses and surfels}
\EndFor
\end{algorithmic}
\label{alg:mad-ba}
\end{algorithm}

\section{Experimental Evaluation}
\label{sec:exp}

\definecolor{Gray}{gray}{0.85}
\begin{table*}[]
\setlength{\tabcolsep}{5pt}
\centering
\begin{tabular}{c!{\color{Gray!40!Gray}\vrule width 1.8pt}l|cc|cccc|cccccc}
\toprule
&  & \multicolumn{2}{c|}{Initial trajectory} & \multicolumn{2}{c}{BALM2 \cite{liu2021balm}} & \multicolumn{2}{c|}{HBA \cite{liu2023hba}} & \multicolumn{2}{c}{Pose-only} & \multicolumn{2}{c}{MAD-\gls{ba} w/o uncert.} & \multicolumn{2}{c}{MAD-\gls{ba} with uncert.} \\ 
& Sequence & \tiny ${\rm RMS}\downarrow$ &  \tiny ${\rm MAX}\downarrow$ &  \tiny ${\rm RMS}\downarrow$ &  \tiny ${\rm MAX}\downarrow$ &  \tiny ${\rm RMS}\downarrow$ &   \tiny${\rm MAX}\downarrow$ &  \tiny ${\rm RMS}\downarrow$ &  \tiny ${\rm MAX}\downarrow$ &  \tiny ${\rm RMS}\downarrow$ &  \tiny ${\rm MAX}\downarrow$ &  \tiny ${\rm RMS}\downarrow$ &  \tiny ${\rm MAX}\downarrow$ \\
\midrule

\multirow{5}{*}{\rotatebox[origin=c]{90}{NC \cite{ramezani2020newer}}}
& quad-easy & 0.098 & 0.218 & 0.093 & 0.249 & 0.131 & 0.801 & 0.086 & 0.214 & 0.086 & 0.183 & \textbf{0.083} & 0.177 \\
& math-easy & 0.090 & 0.189 & 0.162 & 0.420 & 0.135 & 0.606 & 0.087 & 0.189 & 0.054 & 0.163 & \textbf{0.043} & 0.176 \\
& cloister & 0.119 & 0.297 & 0.319 & 1.823 & 0.466 & 2.745 & \textbf{0.098} & 0.364 & 0.112 & 0.356 & 0.106 & 0.318 \\
& stairs & 0.368 & 0.717 & 0.366 & 0.700 & 3.104 & 9.088 & 0.375 & 1.429 & \textbf{0.067} & 0.126 & 0.070 & 0.132 \\
& underground-easy & 0.127 & 0.358 & 0.120 & 0.342 & 0.434 & 1.019 & 0.080 & 0.352 & 0.073 & 0.330 & \textbf{0.072} & 0.331 \\
\arrayrulecolor{black}  %
\midrule
\multirow{6}{*}{\rotatebox[origin=c]{90}{VBR \cite{brizi2024vbr}}}
& Colosseo & 2.920 & 5.873 & 2.877 & 5.553 & 1.123 & 2.763 & 2.891 & 5.887 & 0.698 & 1.585 & \textbf{0.565} & 1.161 \\
& Spagna & 0.963 & 1.912 & 1.058 & 5.479 & 1.628 & 4.625 & 1.053 & 3.953 & 0.412 & 1.229 & \textbf{0.155} & 0.522 \\
& DIAG & 0.341 & 0.804 & 0.425 & 1.945 & 0.960 & 2.510 & 0.335 & 2.014 & \textbf{0.101} & 0.438 & 0.103 & 0.455 \\
& Campus & 1.777 & 4.928 & 2.121 & 6.587 & 1.055 & 3.944 & 1.741 & 5.214 & \textbf{1.025} & 2.972 & 1.109 & 2.973 \\
& Ciampino & 6.078 & 16.097 & 6.086 & 15.669 & 5.641 & 16.645 & 6.040 & 15.881 & 3.878 & 13.711 & \textbf{3.123} & 9.566 \\
& Pincio & 1.488 & 2.914 & 1.719 & 3.293 & 1.508 & 4.589 & 1.499 & 3.304 & \textbf{0.914} & 3.177 & 1.121 & 3.869 \\
\arrayrulecolor{black}  %
\midrule
\multirow{9}{*}{\rotatebox[origin=c]{90}{KITTI \cite{geiger2013vision}}}
& 00 & 5.232 & 11.291 & 5.585 & 16.075 & 4.899 & 9.897 & 5.145 & 11.326 & 3.717 & 7.657 & \textbf{3.544} & 6.473 \\
& 03 & 1.005 & 1.873 & 1.144 & 2.522 & 4.166 & 11.279 & 1.176 & 2.817 & 1.002 & 1.609 & \textbf{0.983} & 1.669 \\
& 04 & 0.444 & 1.162 & 0.507 & 1.226 & 0.532 & 1.776 & 0.399 & 1.075 & \textbf{0.411} & 1.111 & 0.521 & 1.371 \\
& 05 & 3.037 & 10.220 & 3.001 & 9.624 & \textbf{1.321} & 6.083 & 3.006 & 10.290 & 1.335 & 2.497 & 1.426 & 2.682 \\
& 06 & 1.400 & 2.039 & 2.498 & 19.379 & \textbf{0.668} & 1.488 & 1.350 & 2.179 & 0.856 & 1.747 & 0.892 & 1.819 \\
& 07 & 0.767 & 1.192 & 0.748 & 1.197 & \textbf{0.289} & 0.602 & 0.576 & 0.791 & 0.392 & 0.765 & 0.395 & 0.704 \\
& 08 & 4.034 & 12.668 & 8.198 & 67.587 & 4.553 & 12.815 & \textbf{4.027} & 12.319 & 6.535 & 16.982 & 5.449 & 14.873 \\
& 09 & 2.505 & 7.304 & 8.858 & 57.515 & 3.410 & 6.084 & 2.487 & 7.096 & \textbf{1.343} & 2.778 & 1.424 & 2.886 \\
& 10 & 2.024 & 3.720 & 2.725 & 10.738 & 4.179 & 8.107 & 2.051 & 3.807 & \textbf{1.854} & 4.193 & 2.191 & 4.422 \\
\arrayrulecolor{black}  %
\midrule
& avg & 1.741 & 4.289 & 2.430 & 11.396 & 2.010 & 5.373 & 1.725 & 4.525 & 1.243 & 3.181 & \textbf{1.169} & \textbf{2.829} \\
\arrayrulecolor{black}  %
\bottomrule
\end{tabular}
\caption{\textbf{Quantitative results} of ATE for all tested methods across different publicly available datasets. 
Each error is reported in meters, and the last row of the table shows the total average of the error in each column. 
The best results for each sequence (the lowest ${\rm RMSE}$) are
shown in bold.}
\label{tab:ate_evaluation}
\vspace{-4mm}
\end{table*}

In this section, we report the results of our method on different publicly available datasets. 
To run the experiments, we used a PC with an Intel Core i9-9900KF CPU @ 3.60GHz and 64GB of RAM. Our approach consists of a \lidar{} \gls{ba} to refine poses and scene geometry starting from an initialized set of poses and point clouds (\ie~\lidar{} Odometry or SLAM), accompanied by a minimal set of parameters. The results show that our \lidar{} \gls{ba} strategy performs overall better than other existing approaches. Experimental evaluation supports our claim in a wide range of heterogeneous environments, including highways, narrow buildings, stairs, static and dynamic settings.  

We report an extensive experimental campaign on the following datasets: \gls{nc} \cite{ramezani2020newer}, \gls{vbr} \cite{brizi2024vbr} and KITTI \cite{geiger2013vision}. This experimental evaluation includes the comparison with 
\lidar{} global refinement strategies: BALM2 \cite{liu2021balm} and HBA \cite{liu2023hba}. 

\subsection{Datasets}
For the evaluation, we used three publicly available datasets:
\begin{itemize}
\item \gls{nc} \cite{ramezani2020newer}: Collected with a handheld Ouster OS0-128 \lidar{} in structured and vegetated areas. Ground truth was generated using the Leica BLK360 scanner, achieving centimeter-level accuracy over poses and points in the map.
\item \gls{vbr} \cite{brizi2024vbr}: Recorded in Rome using OS1-64 (handheld) and OS0-128 (car-mounted) \lidars{}, capturing large scale urban scenarios with narrow streets and dynamic objects. Ground truth trajectories were obtained by fusing \lidar{}, IMU, and RTK GNSS data, ensuring centimeter-level accuracy over the poses.
\item KITTI \cite{geiger2013vision}: Recorded with a Velodyne HDL-64 in suburban and rural areas of Karlsruhe. While widely used in robotics and computer vision, its GPS/IMU-based ground truth is less accurate than modern high-precision systems, affecting error evaluation in certain sequences.
\end{itemize}

\subsection{Evaluation method}
Since this work focuses on achieving global consistency in both structure and poses, we perform quantitative evaluations using RMS of the \gls{ate} for the $\bbSE(3)$ poses and accuracy, completion, Chamfer-L1 distance, and F-score for the map. The first three parameters used for the map evaluation quantify the geometric discrepancy by measuring the average closest-point distance between the reconstructed map and a reference model, while the F-score measures how accurately they align to each other. For the poses, alignment is first achieved using the Horn method~\cite{horn1988closed}, with timestamps employed to establish pose associations. The RMSE is computed based on the translational differences between all matched poses. We use the three datasets for trajectory evaluation and restrict map evaluation to \gls{nc} dataset, as it is the only one providing a high-resolution ground truth map. Specifically, given the reconstructed map $\mathcal{P}$ and a reference model $\mathcal{Q}$, the accuracy, completion and Chamfer-L1 distance are calculated as follows:

\begin{equation}
\small{
d_{\text{C-L1}}(\mathcal{P}, \mathcal{Q}) = 
\underbrace{\frac{1}{2n} \sum_{p \in \mathcal{P}} \min_{q \in \mathcal{Q}} \| p - q \|}_{\text{accuracy}} + 
\underbrace{\frac{1}{2m} \sum_{q \in \mathcal{Q}} \min_{p \in \mathcal{P}} \| q - p \|}_{\text{completion}}}
\end{equation}
where $m$ and $n$ are the number of points in $\mathcal{P}$ and $\mathcal{Q}$, respectively.

\subsection{Parameters}
All the compared systems require specifying certain parameters that impact their performance. For our method, the most important ones are those responsible for finding correspondences between surfels in the data association step:
\begin{itemize}
    \item  $d_e = 0.5\,\mathrm{m}$: the maximum distance between the points of a surfel and a leaf that are considered for matching;
    \item $d_{n} = 1.0\,\mathrm{m}$: the maximum distance between the points of a surfel and a leaf, calculated along the surfel's normal;
    \item $d_{\theta} = 5$\textdegree: the maximum angle between the normals of a surfel and a leaf;
    \item $\rho_{\mathrm{ker}} = 0.1\,\mathrm{m}$: the threshold value for the Huber robust estimator during the factor graph optimization;
\end{itemize}
and for building the kd-trees:
\begin{itemize}
    \item $b_{\mathrm{max}} = 0.2\,\mathrm{m}$: the maximum size of kd-tree leaves;
    \item $b_{\mathrm{min}} = 0.1\,\mathrm{m}$: the leaf's minimum flatness below which its normal is propagated and assigned to its children.
\end{itemize}
It is important to note that these default values of the parameters were used during the evaluation, as they provide a consistent result across different sequences and datasets. 
In the case of HBA \cite{liu2023hba}, we used the default parameter values, while for BALM2 \cite{liu2021balm}, we changed the initial \textit{voxel size} from $2\,\mathrm{m}$ to $1\,\mathrm{m}$, as it produced better results.

\subsection{Comparison with State-of-the-art}

We select two different approaches to compare our method. The first is BALM2 \cite{liu2021balm}, a global \lidar{} refinement approach that minimizes the distance from feature points to matched edges or planes. Unlike visual SLAM, where features are co-optimized with poses, BALM2 analytically eliminates features from the optimization, reducing computational complexity. However, this approach solely refines poses and does not optimize scene geometry, which limits its ability to handle structural inconsistencies in the map. Its reliance on an adaptive voxelization method for feature correspondence also restricts scalability in highly dynamic or complex environments.

The second approach is HBA \cite{liu2023hba}, designed to address the computational challenges of \lidar{} \gls{ba} on large-scale maps. HBA employs a hierarchical framework that combines bottom-up \gls{ba} with top-down pose graph optimization. While this design improves efficiency by reducing the optimization scale, the reliance on pose graph optimization decouples trajectory estimation from structure refinement, which can lead to inconsistencies in the final map, particularly for long trajectories or complex environments.

In contrast, our method integrates pose and structure optimization within a unified \gls{ba} framework, eliminating the decoupling present in both BALM2 and HBA.

Table~\ref{tab:ate_evaluation} summarizes the quantitative results for trajectory evaluation across all described methods and datasets. The first column shows the error for the initial trajectory, which is required as the starting point for all evaluated systems. We selected the open-source implementation of LOAM \cite{zhang2014loam} for this purpose, as it remains one of the most widely used LiDAR-only odometry and mapping frameworks. However, for KITTI sequences 01 and 02, LOAM failed to produce estimates, so results for these sequences are omitted.

It is important to highlight that the quality of the initial trajectory significantly influences the performance of refinement systems. A poor initial guess can lead to incorrect scan correspondences, causing the optimization to converge to a local minimum. This limitation is evident in the results for BALM2 \cite{liu2021balm}, presented in the second column. BALM2 often shows errors comparable to or worse than those of the initial trajectory, indicating that its voxel-based discretization struggles with the suboptimal starting poses provided by LOAM.

HBA \cite{liu2023hba}, whose results are presented in the third column, performs better on certain sequences such as KITTI 05, 06, and 07, delivering the lowest errors. However, its performance is inconsistent across other datasets, further underscoring its sensitivity to the initial guess. In addition to multi-view \gls{icp}, this approach relies on \gls{pgo} to simplify the optimization. While usually faster, is less accurate compared to \gls{ba}. While parameter tuning could potentially improve the results of both BALM2 and HBA, we maintained a consistent configuration across all sequences to ensure a fair and unbiased evaluation, highlighting the general robustness of each system.

The last three columns in Tab.~\ref{tab:ate_evaluation} present results for different variants of our system. The first variant evaluates our system configured to optimize only the poses while keeping all surfels fixed. The structure of the factor graph remains unchanged, but no adjustments are made to the surfels during optimization.

The next two columns show results for our \gls{ba} method, demonstrating the impact of incorporating uncertainty into the optimization. In the first case, in \eqref{eq:cost-inner}, the uncertainty $\sigma = 1$ assigns equal weight to all measurements. In the second case, $\sigma$ reflects the uncertainty estimated for each measurement during kd-tree creation, as detailed in \secref{sec:lidar_uncertainty}. This strategy gives greater weight to measurement with higher confidence, improving both pose accuracy and map consistency.
As expected, both \gls{ba} configurations significantly reduce errors compared to keeping surfels fixed. Furthermore, incorporating uncertainty into the optimization enhances the overall results by leveraging confidence-weighted measurements.

\begin{figure}[t]
  \centering
  \includegraphics[width=0.99\columnwidth]{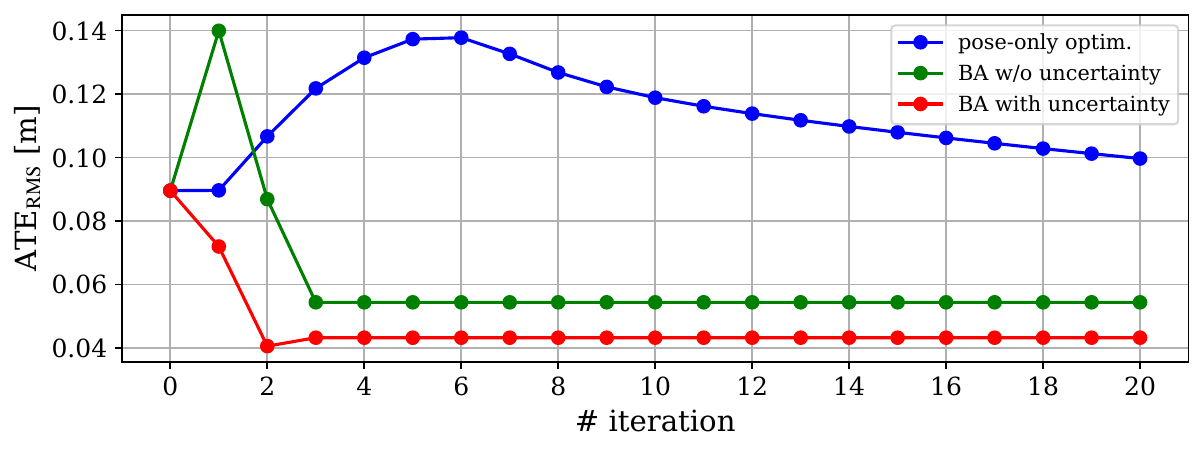}
  \caption{\textbf{ATE for each iteration} of the bundle adjustment algorithm. The plot compares three versions of our system for \textit{math-easy} sequence and shows that both versions of our BA reduce the trajectory error and accelerate the converge of the algorithm related to the pose-only optimization. For BA with uncertainty, the error didn't increase in the first iteration because measurements with higher uncertainty are weighted less during optimization.} 
  \label{fig:math_easy_iterations}
  \vspace{-4mm}
\end{figure}

Interestingly, \figref{fig:math_easy_iterations} shows that considering uncertainty also reduces the number of iterations necessary to achieve convergence, especially compared to pose-only optimization.

\begin{figure}[t]
  \centering
 \includegraphics[width=0.99\columnwidth]{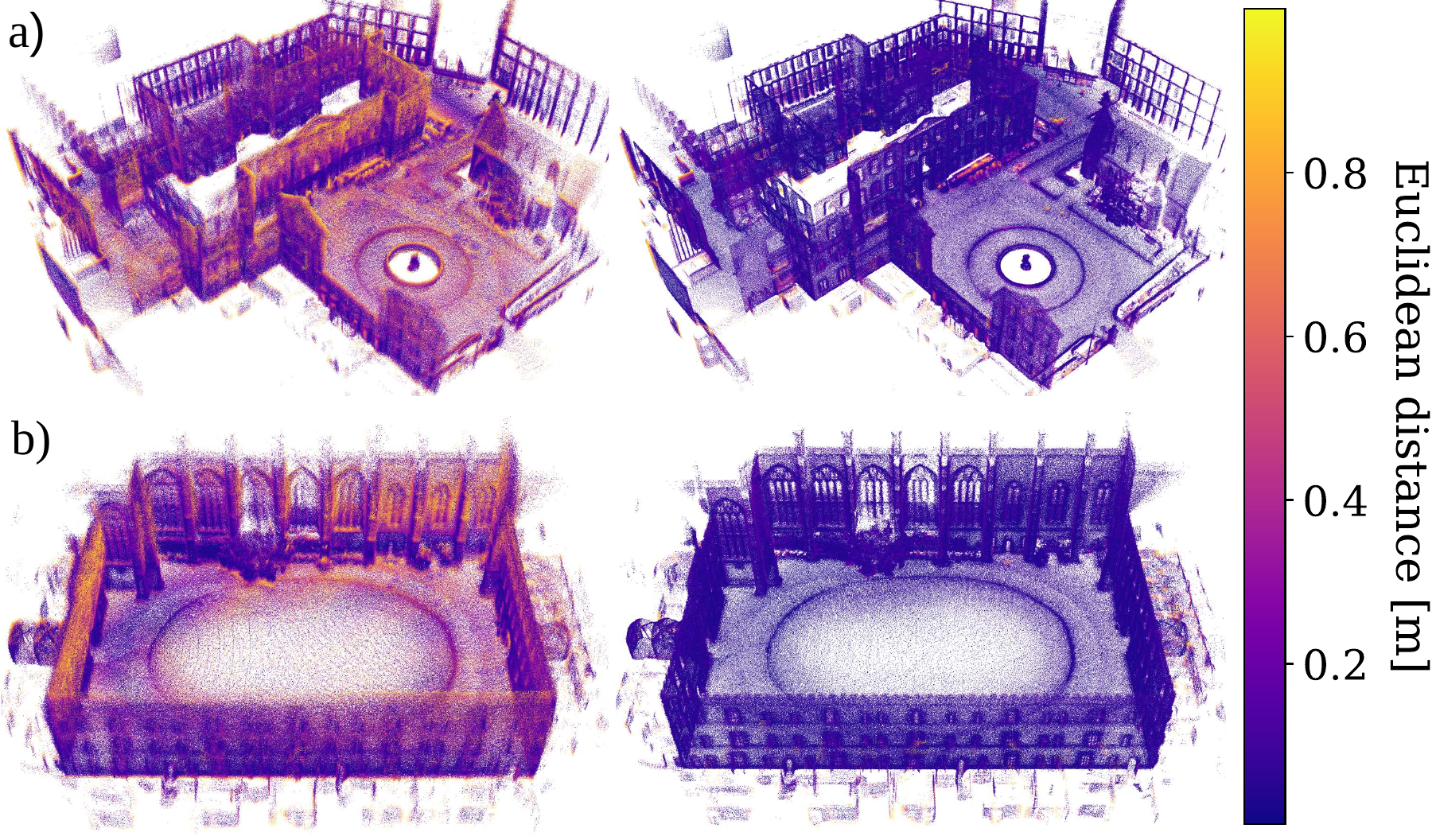}
  \caption{\textbf{Qualitative results} of the maps generated from NC \textit{math-easy} (a) and NC \textit{quad-easy} (b) sequences, where the color of each surfel's point corresponds to its Euclidean distance to the nearest point in the ground truth map. The left-side images are created before the optimization (using the initial trajectory), while the right ones after our BA with uncertainty method.}
  \label{fig:heatmaps}
  \vspace{-2mm}
\end{figure}

\begin{figure}[t]
  \centering
 \includegraphics[width=0.99\columnwidth]{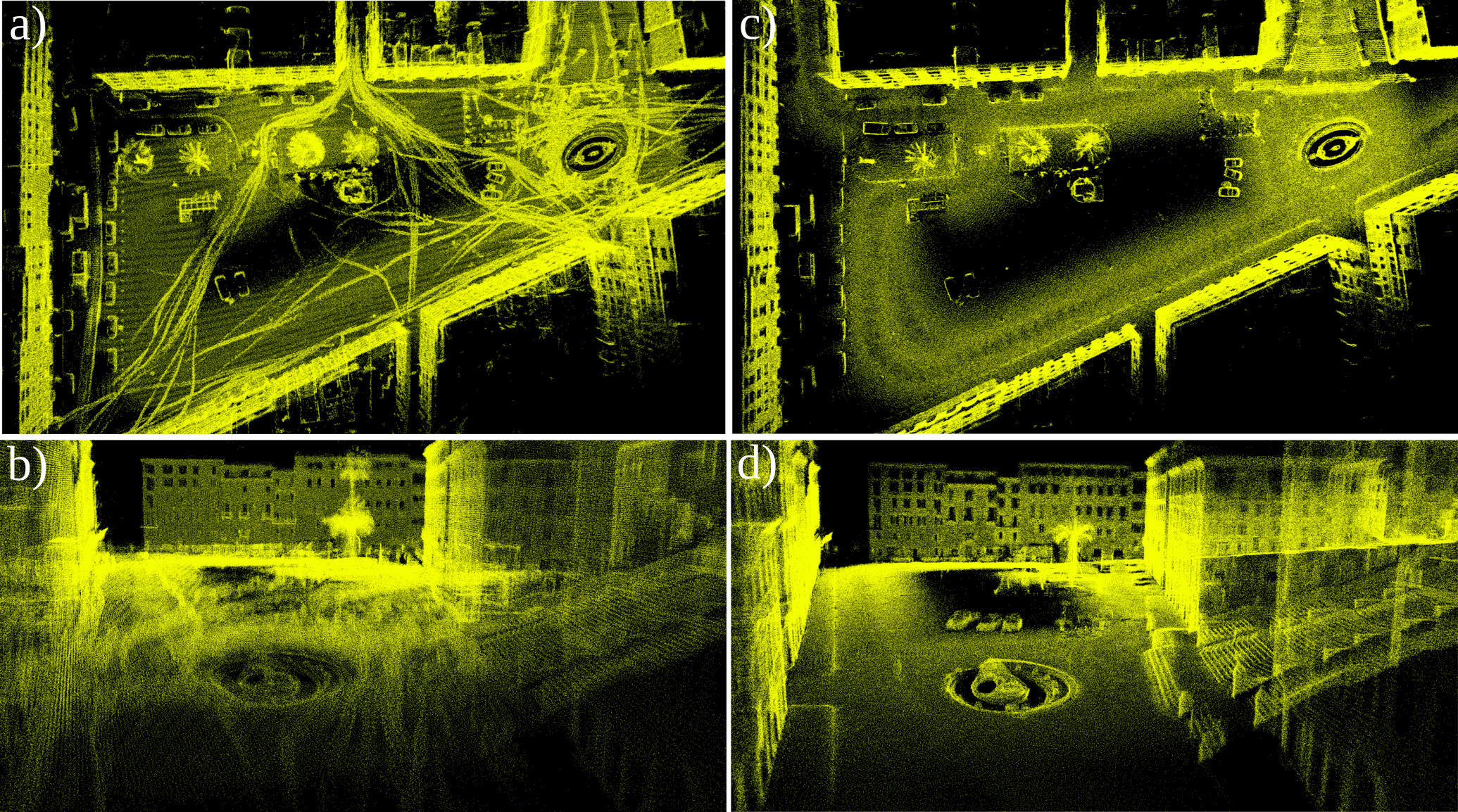}
  \caption{\textbf{Qualitative map results and implicit dynamic filtering.} Map insets generated for VBR \textit{Spagna} sequence by LOAM (a, b) and our MAD-BA method showing the surfel map converted to point cloud (c, d). A benefit of our approach is that it produces a crisp representation of the environment, compensating for noise and filtering points from moving objects, such as cars and people in motion. }
  \label{fig:map_filtering2}
  \vspace{-4mm}              %
\end{figure}

The map evaluation was performed by comparing the ground truth map with the surfel maps generated using the initial trajectory and those generated by MAD-BA with and without uncertainty information, as presented in \tabref{tab:Chamfer-L1}. The table also includes results for the point-based map from the HBA system for comparison. In order to exclude the non-overlapping parts of the refined and reference maps, we specified the threshold for the largest acceptable distance between points in these maps \cite{Deng2023}. The threshold was set to $1\,\mathrm{m}$, however, different values provide analogous results.
Furthermore, we generated a comparison of the maps obtained from \gls{nc} \textit{math-easy} and \textit{quad-easy} sequences, presented in \figref{fig:heatmaps}, which demonstrate a significant improvement in map quality after our bundle adjustment method. Another notable advantage of our map representation is that it implicitly disregards points that were registered on dynamic objects, such as driving cars or walking people. Although these filtering properties are hard to quantify, we present qualitative results of such a case for the part of VBR \textit{ Spagna} sequence in \figref{fig:map_filtering2}.

\begin{table*}[]
\centering
\setlength{\tabcolsep}{4pt}
\begin{tabular}{l|cccc|cccc|cccc|cccc}
\toprule
\multicolumn{1}{l|}{} & \multicolumn{4}{c|}{initial map} & \multicolumn{4}{c|}{HBA} & \multicolumn{4}{c|}{MAD-BA w/o uncertainty} & \multicolumn{4}{c}{MAD-BA with uncertainty} \\
Sequence & acc.$\downarrow$ & comp.$\downarrow$ & C-L1$\downarrow$ & F-sc.$\uparrow$ & acc.$\downarrow$ & comp.$\downarrow$ & C-L1$\downarrow$ & F-sc.$\uparrow$ & acc.$\downarrow$ & comp.$\downarrow$ & C-L1$\downarrow$ & F-sc.$\uparrow$ & acc.$\downarrow$ & comp.$\downarrow$ & C-L1$\downarrow$ & F-sc.$\uparrow$ \\
\midrule
quad-easy & 16.39 & 22.52 & 19.45 & 68.35 & 14.59 & 42.93 & 28.76 & 41.41 & 12.72 & 6.97 & 9.85 & 92.04 & 12.70 & 6.93 & \textbf{9.82} & \textbf{92.07} \\
math-easy & 19.22 & 26.87 & 23.05 & 58.55 & 16.86 & 43.49 & 30.17 & 38.70 & 14.44 & 10.65 & 12.55 & 87.20 & 14.35 & 10.54 & \textbf{12.44} & \textbf{87.26} \\
cloister & 22.78 & 16.31 & 19.54 & 69.18 & 18.62 & 39.18 & 28.90 & 45.76 & 21.78 & 7.34 & 14.56 & 77.86 & 21.18 & 6.95 & \textbf{14.06} & \textbf{78.96} \\
stairs & 28.91 & 23.66 & 26.29 & 55.51 & 33.82 & 32.35 & 33.08 & 41.41 & 24.32 & 7.22 & \textbf{15.77} & \textbf{73.50} & 24.56 & 7.28 & 15.92 & 73.36 \\
\bottomrule
\end{tabular}
\caption{
\textbf{Quantitative results of the map evaluation.} The comparison includes the following metrics: accuracy (acc.), completion (comp.), Chamfer-L1 (C-L1), and F-score (F-sc.). The first three metrics are given in centimeters, while the F-score is expressed as a percentage. The column \textit{initial map} presents results for surfel map created using the initial trajectory before applying our \lidar{} Bundle Adjustment method. A threshold of $1\,\mathrm{m}$ was used to exclude non-overlapping regions of the map, and an additional threshold of $0.2\,\mathrm{m}$ was used to calculate F-score. The \textit{underground-easy} is not included in this table, as there is no ground truth map for this sequence.}
\label{tab:Chamfer-L1}
\vspace{-4mm}
\end{table*}

\begin{figure}[h!]
\centering
\includegraphics[width=1.0\columnwidth]{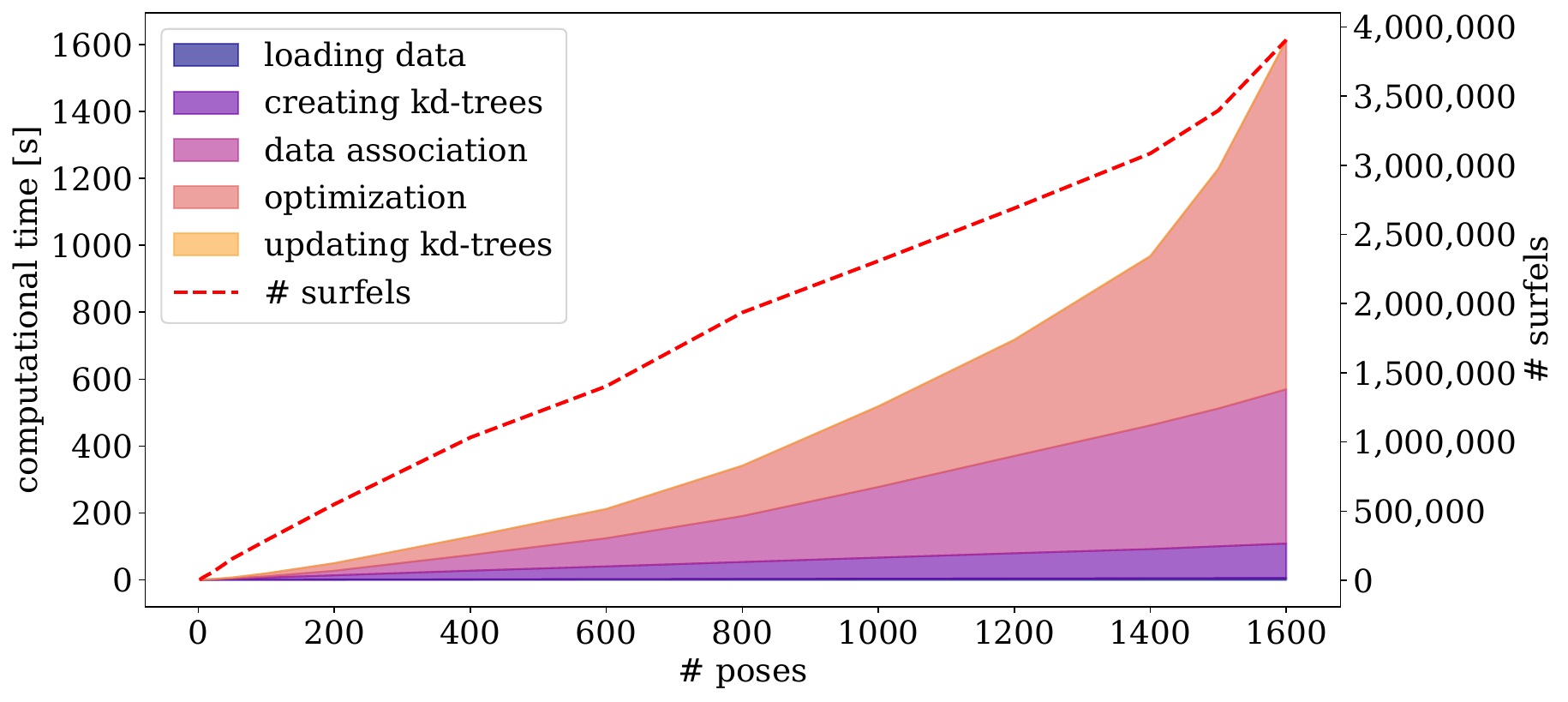}
\caption{\textbf{Computation time versus the number of poses.} The optimization process exhibits exponential time complexity, while the data association scales quadratically, representing the most computationally intensive components.}
\label{fig:computational_time}
\vspace{-4mm}
\end{figure}

To provide a deeper understanding of the computational requirements of the system, we also evaluated the time for each major component separately, as shown in \figref{fig:computational_time}.

\subsection{Ablation study}
To evaluate the impact of the point cloud density on performance, we downsampled the LiDAR scans to 16 lines to simulate the VLP-16 sensor. The resulting $\mathrm{ATE}$ values for this configuration are provided in the \tabref{tab:sim_vlp16}, illustrating the accuracy of the system with reduced sensor data. Results are presented for selected sequences of the \gls{nc} and \gls{vbr} datasets, showing that our method performs effectively with sparse point clouds, except in vertical motion scenarios such as \textit{DIAG} and \textit{stairs}. In these cases, the limited vertical field of view significantly degrades performance due to insufficient points captured on the floor or ceiling.

\begin{table}[h]
\centering
\begin{tabular}
{l|cc|cc}
\toprule
 & \multicolumn{2}{c|}{MAD-BA (full cloud)} & \multicolumn{2}{c}{MAD-BA (downsampled)} \\
Sequence & \tiny ${\rm RMS}\downarrow$ & \tiny ${\rm MAX}\downarrow$ & \tiny ${\rm RMS}\downarrow$ & \tiny ${\rm MAX}\downarrow$ \\
\midrule
quad-easy & 0.083 & 0.177 & 0.093 & 0.223 \\
math-easy & 0.043 & 0.176 & 0.066 & 0.199 \\
stairs & 0.070 & 0.132 & {\bf 2.179} & {\bf 4.108} \\
\midrule
Colosseo & 0.565 & 1.161 & 0.748 & 2.507 \\
Ciampino & 3.123 & 9.566 & 4.133 & 11.817 \\
DIAG & 0.103 & 0.455 & {\bf 1.050} & {\bf 4.337} \\
\bottomrule
\end{tabular}
\caption{\textbf{ATE results for downsampled LiDAR scans}. It simulates VLP-16. Results from \tabref{tab:ate_evaluation} are shown for comparison and the largest differences are bolded.}
\label{tab:sim_vlp16}
\vspace{-4mm}
\end{table}

\begin{figure}[h]
\vspace{-2mm}
\centering
\includegraphics[width=0.90\columnwidth]{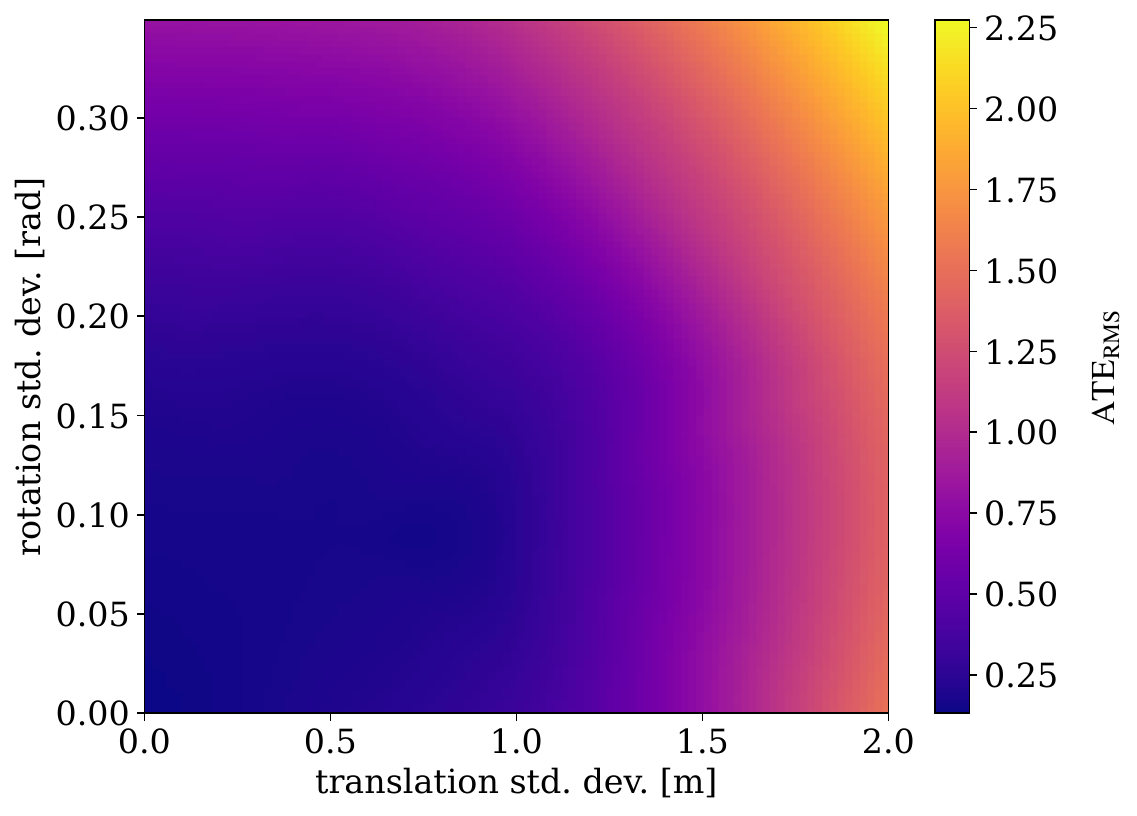}
\caption{\textbf{Rubustenss of MAD-BA}. The figure shows the ATE as a function of perturbation (Gaussian noise with zero mean) added to the initial trajectory.}
\label{fig:perturbation_heatmap}
\vspace{-4mm}
\end{figure}

We evaluated also the robustness of the system to initial trajectory errors by injecting Gaussian noise at varying intensities. Based on \figref{fig:perturbation_heatmap}, it can be seen that the method reliably converges to an optimal solution with a standard deviation of the initial noise of up to around $1\,\mathrm{m}$ for translation and $0.2\,\mathrm{rad}$ for rotation, demonstrating a strong initialization tolerance.

\section{Conclusion}
\label{sec:conclusion}

This work introduces a novel framework for simultaneous pose and 3D map optimization, using a surfel-based map representation. We also propose a generalized \lidar{} uncertainty model for accurate weighting in optimization. Experimental results on multiple public datasets show that our system surpasses existing LiDAR-based methods in robustness and accuracy. It precisely estimates poses and maps while mitigating dynamic artifacts and \lidar{} skewing, producing clean, high-fidelity 3D maps.

Our open-source implementation aims to benefit the community. Future work will enhance scalability for larger environments, integrate multi-modal sensor data (e.g., cameras and inertial sensors), and extend the framework to highly dynamic settings. Additionally, we plan to explore real-time implementations to bridge batch optimization with practical deployment in robotics and autonomous systems.

\bibliographystyle{IEEEtran}

\balance
\bibliography{citations}
\end{document}